\documentclass[journal]{ieeetran}
\usepackage{cite}
\usepackage{graphicx}
\usepackage{epstopdf}
\usepackage{amsmath}
\usepackage{cases}
\usepackage[caption=false,font=footnotesize]{subfig}
\usepackage{fixltx2e}
\usepackage{amssymb}
\usepackage{mathtools}
\usepackage{bm}
\usepackage{multirow}
\usepackage{color}
\usepackage{algorithm}
\usepackage{algorithmic}
\usepackage{amsthm}
\usepackage{mathrsfs}
\usepackage{subfig}

\usepackage{textcomp}

\newcommand{\tabincell}[2]{\renewcommand\arraystretch{0.9}\begin{tabular}{@{}#1@{}}#2\end{tabular}}

\begin{document}


\title{Achieving an Accurate Random Process Model for PV Power using Cheap Data: Leveraging the SDE and Public Weather Reports\vspace{-1pt}}


\author{Yiwei~Qiu, 
Jin~Lin, 
Zhipeng~Zhou, 
Ningyi~Dai, 
Feng~Liu, and 
Yonghua~Song 

\thanks{Financial support came from the National Key R\&D Program of China (2018YFB0905200), National Natural Science Foundation of China (51907099), Beijing Nova Program of Science and Technology (Z191100001119073), Science and Technology Development Fund, Macau SAR (SKL-IOTSC-2021-2023). (Corresponding author: Jin Lin)}
\thanks{Y. Qiu is with College of Electrical Engineering, Sichuan University, Chengdu, 610065, China. 
}
\thanks{J. Lin, and F. Liu are with the State Key Laboratory of the Control and Simulation of Power Systems and Generation Equipment, Tsinghua University, Beijing, 100087, China. 
}
\thanks{Z. Zhou and N. Dai are with the State Key Laboratory of Internet of Things for Smart City, University of Macau, Macau 999078, China.}%
\thanks{Y. Song is with the State Key Laboratory of Internet of Things for Smart City, University of Macau, Macau 999078, China, and also with the Department of Electrical Engineering, Tsinghua University, Beijing 100087, China. 
}
}

\maketitle

\begin{abstract}
  The stochastic differential equation (SDE)-based random process models of volatile renewable energy sources (RESs) jointly capture the evolving probability distribution and temporal correlation in continuous time. It has enabled recent studies to remarkably improve the performance of power system dynamic uncertainty quantification and optimization. However, considering the non-homogeneous random process nature of PV, there still remains a challenging question: how can a realistic and accurate SDE model for PV power be obtained that reflects its weather-dependent uncertainty in online operation, especially when high-resolution numerical weather prediction (NWP) is unavailable for many distributed plants? To fill this gap, this article finds that an accurate SDE model for PV power can be constructed by only using the cheap data from low-resolution public weather reports. Specifically, an hourly parameterized Jacobi diffusion process is constructed to recreate the temporal patterns of PV volatility during a day. Its parameters are mapped from the public weather report using an ensemble of extreme learning machines (ELMs) to reflect the varying weather conditions. The SDE model jointly captures intraday and intrahour volatility. Statistical examination based on real-world data collected in Macau shows the proposed approach outperforms a selection of state-of-the-art deep learning-based time-series forecast methods.
\end{abstract}


\section{Introduction}
\label{sec:intro}

\bstctlcite{BSTcontrol}

\IEEEPARstart{I}{nstalled} capacities of distributed photovoltaic (PV) stations across the globe continue to surge. As of the end of 2019, worldwide installations of distributed PV reached $251.6$ GW, $40.39\%$ of the total PV installations \cite{Trends2020}, with many mounted on roofs or integrated into residential, commercial or industrial buildings \cite{fina2020cost}. 
To cope with its volatile and intermittent nature
in power system regulation and energy management, 
knowing the underlying random process of PV generation could be beneficial.


As a typical renewable energy source (RES), PV generation has strong weather-dependent uncertainties. Its probability distribution and temporal correlation vary as the meteorological conditions change. To capture these uncertainty characteristics in continuous time to improve the performance of power system operation and control, \emph{stochastic differential equation (SDE)}-based random process models \cite{lingohr2019stochastic,iversen2014probabilistic,badosa2017day} have been proposed.

The SDE is typically a differential equation driven by the \emph{Brownian motion} \cite{Pardoux2014Stochastic}.
It has been accepted by the community for modeling RESs, such as wind \cite{
jonsdottir2019data}, 
solar \cite{lingohr2019stochastic,iversen2014probabilistic,badosa2017day}, and tidal power \cite{jonsdottir2019modeling}, in recent years.
Because the SDE is a continuous-time random process, it captures the random volatilities of RESs at very high resolutions.
In fact, based on the renowned stochastic calculus theory \cite{Pardoux2014Stochastic} that dominates the finance industry \cite{wu2016dynamic},
recent studies in renewable energy engineering have already taken advantage of these SDE models in power system uncertainty quantification (UQ) \cite{wang2015long,Chen2018Stochastic,qiu2021nonintrusive,qiu2020fast} and stochastic optimization (SO) \cite{
qin2019stochastic,Li2019Compressive,chen2019optimal,Chen2020OptimalC,qiu2020stochastic}. 

For instance, Chen et al. \cite{Chen2018Stochastic} and Qiu et al. \cite{qiu2021nonintrusive} derived analytical approximations to the statistical performances for power system AGC and transient stability evaluation under 
renewable volatility based on the SDE. 
The former one was then used to construct a stochastic model predictive controller (MPC) for power system AGC \cite{chen2019optimal} 
and the energy management of ADNs \cite{Chen2020OptimalC}.
The complexity of these stochastic optimizations is almost the same as a deterministic MPC, 
which proves to be far more efficient than traditional scenario-based stochastic evaluation and optimization approaches
\cite{chen2019optimal,
Chen2020OptimalC}.

However, there is still one key but challenging question not addressed by previous works: \emph{How can a realistic SDE model for PV be obtained considering its weather-dependent uncertainty in real-world online operation?}
In \cite{
qin2019stochastic,Li2019Compressive}, the Ornstein-Uhlenbeck process 
was employed to model the 
PV power, 
but 
its time-constant statistics do not comply with the varying weather conditions, and its unbounded Gaussian distribution is unrealistic. In fact, Lai et al. \cite{lai2018modeling} pointed out that the traditional autoregressive (AR) or Gaussian process models may fail to model PV power because they cannot capture the mixture of long-term and short-term patterns. In \cite{Chen2018Stochastic,chen2019optimal,Chen2020OptimalC}, non-Gaussian SDE models were adopted, but the SDE models themselves are based on historical statistics and cannot reflect the varying weather in the real-world operation.

Data-based random process models have been proposed to capture the statistics of PV power on a seasonal or yearly timescale \cite{lingohr2019stochastic,ekstrom2017statistical
}. 
These models are very useful in energy planning \cite{ekstrom2017statistical} 
and financial applications such as PV futures pricing \cite{lingohr2019stochastic}. However, because these models only fit historical data, they cannot be used for online operation given the current meteorological conditions.

To meet the requirement of power system online operation, SDE models obtained by likelihood regression of high-resolution numerical weather prediction (NWP) 
\cite{iversen2014probabilistic,badosa2017day} were also proposed. 
Unfortunately, high-resolution NWPs are unavailable for many distributed PV stations, such as the rooftop and building-integrated ones. Alternatively, if the low-resolution but widely available public weather report can be used to build an SDE model for PV power without losing too much accuracy,
it could be of great benefit.

\begin{figure}[tb]
  \centering
  \includegraphics[width=3.45in]{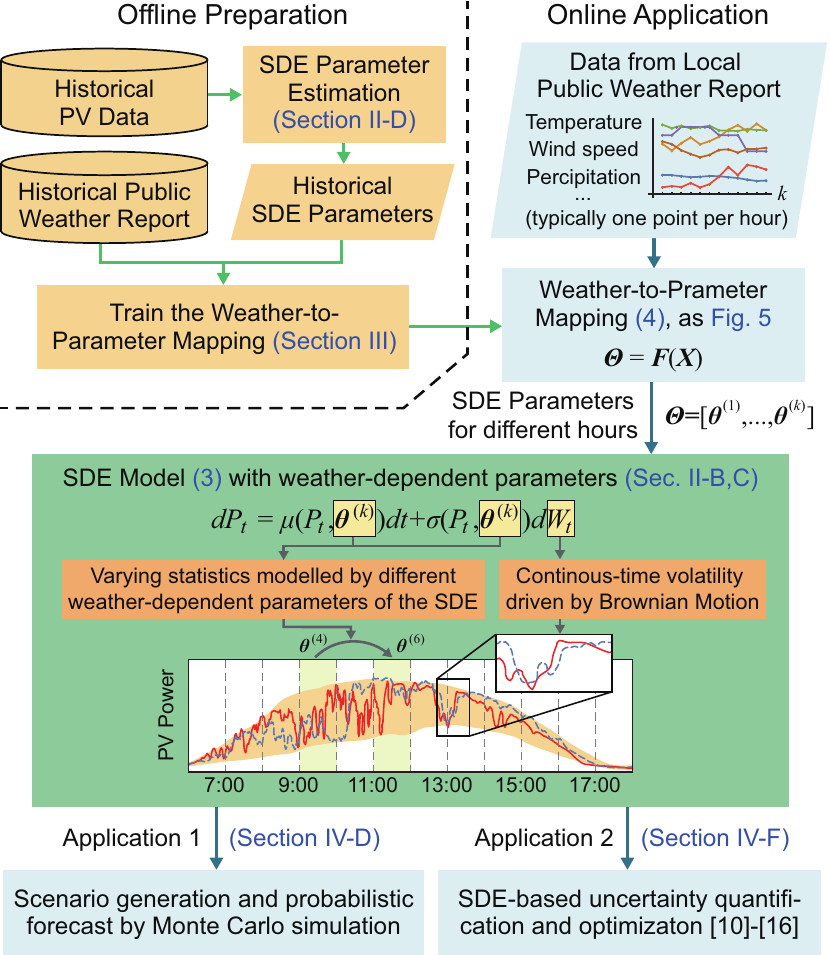}\vspace{-4pt}
\caption{Conceptual framework of this work.}\vspace{-10pt}
  \label{fig:frame}
\end{figure}

To fill the gap, this work finds that an accurate weather-dependent SDE model for PV power can be constructed using only widely available public weather reports. 
Specifically, a Jacobi diffusion process 
is built to recreate the continuous-time pattern of PV fluctuation. To capture the weather-dependent volatility and to accommodate the hourly resolution of the public weather report, the model is parameterized hourly. Transitions between clear sky, cloudy, and rainy conditions can be represented by only changing the parameters.
Finally, a mapping from public weather reports to these parameters is established by an ELM ensemble. 
Therefore, given the public weather report of the operation day, an accurate SDE model for PV generation can be directly obtained.
The conceptual framework of this work is summarized in Fig. \ref{fig:frame}.

On the other hand, extensive researches have been made on weather-dependent scenario generation \cite{camal2019scenario,chen2018model} and forecast \cite{
mashlakov2021assessing,wang2018adaptive}, providing trajectories and statistical information like interval and distribution.
Especially, the latest deep learning (DL)-based time-series forecasts capture the non-uniform patterns of PV power \cite{salinas2020deepar,lai2018modeling,chen2020probabilistic,dabrowski2020forecastnet}. Mashlakov et al. \cite{mashlakov2021assessing} gave a comprehensive review of this topic.
Nevertheless, compared to the proposed SDE-based model, DL-based forecasts have two shortcomings.
First, DL-based forecasts only provide scenarios for power system UQ and SO, which are time-consuming. In contrast, SDE-based UQ and SO are very efficient \cite{wang2015long,Chen2018Stochastic,qiu2021nonintrusive,qin2019stochastic,Li2019Compressive,chen2019optimal,Chen2020OptimalC}.  
Second, the patterns of PV power are entirely learned from data. In contrast, the SDE-based model embeds the patterns as prior knowledge. Hence, with limited information in the public weather report, the proposed model achieves a better forecast than the DL-based ones. See discussions in Section \ref{sec:dl}.

Highlights of this work include:
\begin{enumerate}
\item A weather-dependent continuous random process model for PV generation is first proposed in the form of an SDE. Fast continuous-time fluctuations and 
statistical changes under varying weather conditions are jointly captured.

\item The model only relies on a low-resolution public weather report that is, in general, freely accessible from the local meteorological service. 

\item Compared to a selection of deep learning-based time-series forecasts, the proposed SDE model is more accurate 
    when only 
    public weather reports are available.

\item  The proposed model exclusively enables the latest SDE-based stochastic evaluation and optimization methods, far more efficient than the traditional scenario-based ones.
\end{enumerate}

The remainder of this article is as follows. Section \ref{sec:sde} constructs the weather-dependent SDE model for PV generation. In Section \ref{sec:map}, the SDE parameters of historical PV generation are estimated, and the weather-to-parameter mapping is established. Finally in Section \ref{sec:case}, the proposed model is validated via statistical similarity examination and comparison with a selection of deep learning-based forecast methods.


\section{Modeling the Weather-Dependent Random Process of PV Power with an SDE}
\label{sec:sde}

\subsection{Normalization of PV Data}
\label{sec:normal}

To characterize the random process of PV with respect to only the weather conditions, the 
deterministic pattern of 
the Earth's movement is removed. Instead, the normalized PV power defined as follows will be modeled by the SDE:
\begin{align}
  P_t = {P_t^0}/{\big(P^{\text{rate}}\cos(\alpha_t)\big)} \label{eq:norm}
\end{align}
\noindent
where $P_t$ is the normalized PV power; $P_t^0$ is the actual PV power at time $t$; $P^{\text{rate}}$ is the rating of the station; and $\alpha_t$ is the apparent sun position angle, 
a deterministic function of time.

\begin{figure}[tb]
  \centering
  \includegraphics[scale=0.91]{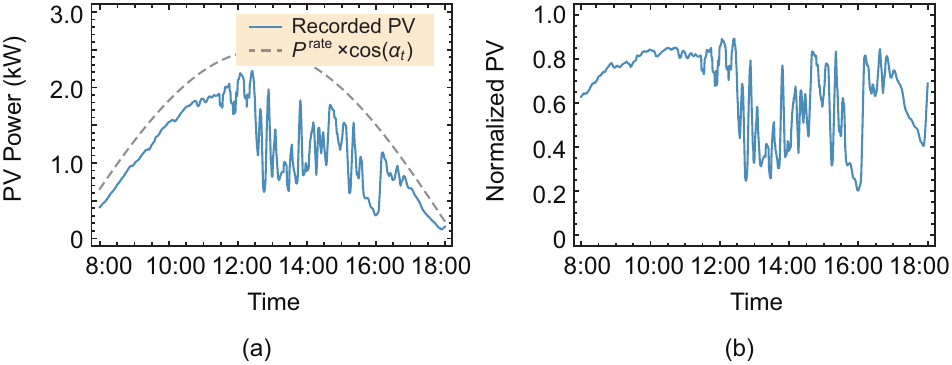}\vspace{-5pt}
\caption{PV power of a rooftop station at the University of Macau on Jun 10, 2018 and its normalization. (a) Actual PV power. (b) Normalization.}
  \label{fig:normalization}
\end{figure}

For example in Fig. \ref{fig:normalization},
the normalization removes the deterministic pattern of the Earth's movement.
The clear sky in the morning 
and cloud shadowing in the afternoon 
are clearly seen. 
Therefore, we can focus on the impact of the weather conditions on the uncertainty. 
Without ambiguity, this article refers to the normalized PV power simply as PV power.

\subsection{SDE Model of PV Considering Various Weather Conditions}
\label{sec:jacobi}

PV power varies within a finite interval bounded by clear sky irradiation, and statistical analysis shows that PV power has an approximately exponential autocorrelation on the ultrashort-term timescale \cite{qiu2021nonintrusive}. To capture these characteristics in continuous time, we construct a parameterized \emph{Jacobi diffusion process} \cite{Pardoux2014Stochastic} to model PV power, an SDE formulated by
\begin{align}
  dP_t & = a (b - P_t)dt +  \sqrt{\beta (P_t - c)(d - P_t)} dW_t  \nonumber \\
  & \triangleq \mu \big( P_t; \bm{\theta} )dt + \sigma \big( P_t; \bm{\theta} ) dW_t,  \label{eq:ito}
\end{align}
\noindent
where $W_t$ is the \emph{Brownian motion} \cite{Pardoux2014Stochastic}; $t$ is time in seconds; $\mu(\cdot;\cdot)$ and $\sigma(\cdot;\cdot)$ are named the \emph{drift} and \emph{diffusion terms}, respectively; and $\bm{\theta} \triangleq [ a, b, \beta, c, d ]^{\mathrm{T}}$ is the parameter vector.

\begin{figure}[tb]
  \centering
  \includegraphics[scale=0.91]{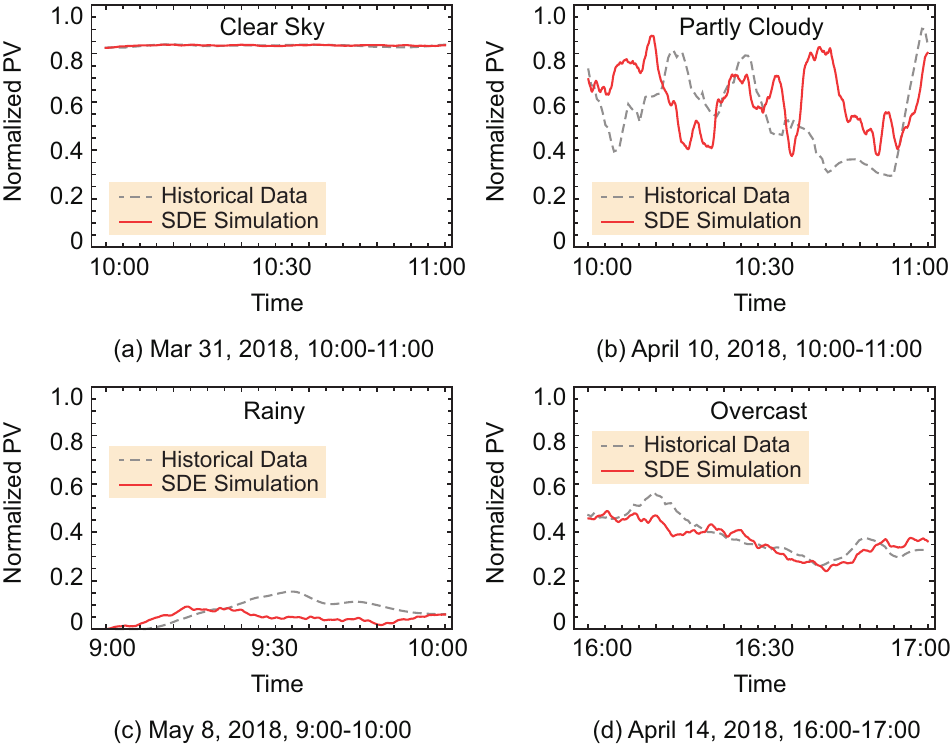}\vspace{-5pt}
\caption{The normalized PV power for one hour under different weather conditions and the simulation trajectory of the SDE model.}
  \label{fig:hourPV}\vspace{-11pt}
\end{figure}

By changing the parameters, the SDE model (\ref{eq:ito}) transits between different meteorological conditions. For instance, in Fig. \ref{fig:hourPV}, simulation samples of the SDE model under four different weather conditions, i.e., clear sky, cloudy, rain, and overcast, are compared to the actual PV data. The public weather report data 
and the corresponding SDE parameters are given in Table \ref{tab:hourPV}. Here the parameters are identified from the actual PV data using the method presented in Section \ref{sec:estimation}.

The intuitive meanings of the parameters are explained as follows. The drift term $\mu(\cdot)$ determines the mean-reverting property of $P_t$, where $b$ is the reverting target that determines the overall level of PV output, and $b$ is the reverting velocity that determines the time constant of the autocorrelation; i.e., $P_t$ has an exponential autocorrelation with time constant $1/{a}$.

\begin{table}[tb]\scriptsize \vspace{-4pt}
  \renewcommand{\arraystretch}{1.26}\vspace{-3pt}
  \caption{Different Weather Conditions and the Corresponding SDE Parameters for PV Generation during Typical Hours}\vspace{-4.5pt}
  \label{tab:hourPV}
  \centering
  \begin{tabular}{ccccc}
  \hline \hline
  \multicolumn{5}{c}{Typical Hours with Different Weather Conditions} \\
  \hline
  Date              & 3/31/2018     & 4/10/2018     & 5/8/2018      & 4/14/2018   \\
  Hour              & 10:00-11:00   & 10:00-11:00   & 9:00-10:00    & 16:00-17:00 \\
  Weather Type      & Clear Sky     & Partly Cloudy & Rainy         & Overcast   \\
  Figure            & \ref{fig:hourPV}(a)   & \ref{fig:hourPV}(b)   & \ref{fig:hourPV}(c)   & \ref{fig:hourPV}(d) \\ \hline
  \multicolumn{5}{c}{Weather Conditions}                    \\ \hline
  Temperature (\textcelsius{})
                    & 24.14         & 24.20         & 23.27         & 25.85    \\
  Wind Speed (m/s)  & 11.298     & 7.044         & 9.072         & 9.294 \\
  Wind Direction    & E             & E             & NE            & S \\
  Humidity ($\%$)   & 57            & 75            & 98            & 87    \\
  Precipitation (mm)& 0             & 0             & 7.8           & 0 \\
  Cloud (Okta)      & 6             & 6             & 9             & 9  \\
  ATM (hPa)         & 1004.4        & 1003.7        & 996.0         & 996.6 \\
  Irradiance (MJ/m$^\text{2}$)
                    & 2.27          & 2.39          & 0.02          & 0.67  \\ \hline
  \multicolumn{5}{c}{Parameters of the SDE Model (\ref{eq:ito}) for the Corresponding Hour}            \\ \hline
  $a$               & 0.3298        & 0.2095        & 0.0760        & 0.0461 \\
  $b$               & 0.8333        & 0.5496        & 0.0519        & 0.3547  \\
  $\beta$           & 0.0348        & 0.1946        & 0.0519        & 0.1064  \\
  $c$               & 0.6895        & 0.1263        & 0             & 0.2267  \\
  $d$               & 0.8477        & 0.9930        & 0.3143        & 0.6209    \\
  \hline \hline
  \end{tabular}
\end{table}

On the other hand, the diffusion term $\sigma(\cdot)$ determines the volatility of $P_t$. Parameter $\beta$ quantifies the intensity of the volatility, as clearly a larger $\beta$ means a larger impact from the Brownian motion. Parameters $c$ and $d$ determine the bounds of $P_t$; i.e., $P_t \in [c, d]$. This is because when $P_t$ approaches $c$ or $d$, the diffusion term approaches zero, and $P_t$ is reverted towards $b$ dominantly driven by the drift term.

For instance, under the clear sky in Fig. \ref{fig:hourPV}(a), $b$ is large to represent a high mean value of PV output, and a small $\beta$ depicts the mild volatility. Comparatively, under the cloudy condition in Fig. \ref{fig:hourPV}(b), the smaller $b$ represents the smaller mean value, the large distance between $c$ and $d$ depicts the large interval where $P_t$ varies, and $\beta$ is large to depict the intensive volatility. Moreover, when wind velocity is high, $a$ becomes larger to model the rapid fluctuations of the PV output.

As can be seen, the temporal pattern of PV power volatility is already embedded in the SDE model (\ref{eq:ito}) as prior knowledge and properly parametrized. 
Unlike deep learning-based models, we do not need to learn the temporal patterns from data.


\subsection{Piecewisely Parameterized SDE Model for PV Power under Changing Weather Conditions}
\label{sec:piecewise}

In order to consider the impact of changing weather conditions on the random fluctuations of PV power and to adapt to the public weather report data with hourly resolution, we parameterize the SDE (\ref{eq:ito}) on an hourly basis, as follows:
\begin{align}
  dP_t & = a^{(k)} (b^{(k)} - P_t)dt +  \sqrt{\beta^{(k)} (P_t - c^{(k)})(d^{(k)} - P_t)} dW_t  \nonumber \\
  & \triangleq \mu \big( P_t; \bm{\theta}^{(k)} )dt + \sigma \big( P_t; \bm{\theta}^{(k)} ) dW_t, \ k = \lceil t / 3600 \rceil,  \label{eq:itohour}
\end{align}
\noindent
where superscript $^{(k)}$ represents the parameter of the $k$th hour; and $ \lceil \cdot \rceil$ represents round up.

The parameters of hour $k$ are determined by the corresponding meteorological conditions. As the hour index $k$ increases, the SDE parameters change, but $P_t$ remains continuous. 
Furthermore, with $a^{(k)}>0$ the SDE is mean-reverting. This means its simulation trajectories do not diverge. 
In this manner, the intraday and intrahour random processes of PV with weather-dependent volatility are jointly characterized.

\begin{figure}[tb]
  \centering
  \includegraphics[scale=0.91]{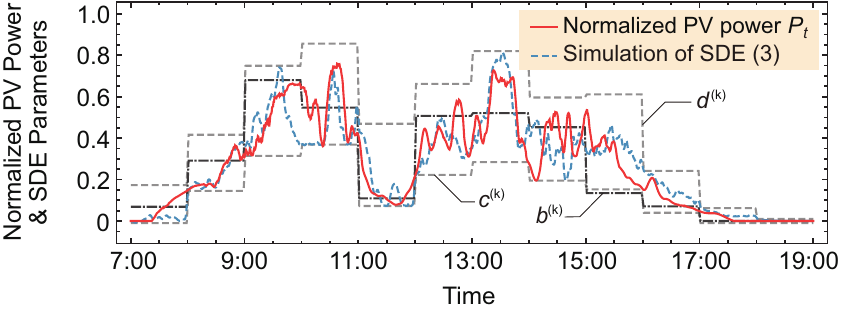}\vspace{-5pt}
\caption{Normalized PV power on Jan 23, 2018, the SDE parameters, and a simulation trajectory of the SDE model.}
  \label{fig:paraJan23}\vspace{-12pt}
\end{figure}

For example, Fig. \ref{fig:paraJan23} shows the normalized PV power $P_t$ of the rooftop station at the University of Macau on Jan 23, 2018, the SDE parameters $b^{(k)}$, $c^{(k)}$, and $d^{(k)}$, and a simulation of the SDE model (\ref{eq:itohour}). The proposed SDE model (\ref{eq:itohour}) realistically captures the varying volatility in continuous time.

Notably, \emph{the SDE parameters are quantitatively related to the meteorological data in the public weather report}, including temperature, wind velocity, 
etc. Although public weather reports only provide these data with an hourly resolution, 
they do statistically determines the continues-time volatility of PV power.
Aided by the temporal patterns embedded in the SDE model (\ref{eq:itohour}), the intra-hour volatility can be accurately captured despite the low resolution of the public weather report.

This means if we can establish a multi-input-multi-output (MIMO) function that maps the public weather report to the SDE parameters, a realistic weather-dependent random process model for PV, i.e., SDE (\ref{eq:itohour}), can be obtained.

For convenience, we denote such an MIMO function as
\begin{align}
  \bm{\Theta} = \bm{F} (\bm{X}), \label{eq:map}
\end{align}
\noindent
where $\bm{\Theta} \triangleq [\bm{\theta}^{(1)}, \ldots, \bm{\theta}^{(m)}]$ is the SDE parameter vector for a day; $\bm{X} \triangleq [\bm{x}^{(1)}, \ldots, \bm{x}^{(m)}]$ is the vector of the public weather report data; $m$ is the number of daytime hours.

Supposing $\bm{F}(\cdot)$ is known, given the daily public weather report data, 
the corresponding SDE model for PV power is directly obtained. 
However, although the SDE parameters and weather data are somewhat related, 
the quantitative relation is not obvious and easy to find. Therefore, finding a decent approximation of $\bm{F}(\cdot)$ 
is the remaining work of this article.

\section{Mapping Weather Reports to SDE Parameters}
\label{sec:map}

\subsection{Input-Output Definition}
\label{sec:inputoutput}

The remaining task is to find an approximation of $\bm{\Theta} = \bm{F} (\bm{X})$, the weather-to-parameter mapping. Specifically, the output $\bm{\Theta}$ is the parameters of SDE (\ref{eq:itohour}) for $m$ hours; i.e., $\bm{\Theta} \triangleq [\bm{\theta}^{(1)}, \ldots, \bm{\theta}^{(m)}]$, where $\bm{\theta}^{(i)} = [a^{(i)}, b^{(i)}, \beta^{(i)}, c^{(i)}, d^{(i)}]$ represents the parameters for the $i$th hour. The input $\bm{X} \triangleq [\bm{x}^{(1)}, \ldots, \bm{x}^{(m)}]$ is the public weather report data of the $m$ hours. Each $\bm{x}^{(i)}$ consists of temperature (in \textcelsius), precipitation amount (in mm), atmospheric pressure (in hPa), wind velocity (in m/s) and direction, solar irradiance (in MJ/m$^{2}$), insulation duration (in h), cloud amount (in Okta), etc., of hour $i$. 

\subsection{Data Preparation: Parameter Identification for Historical PV Generation}
\label{sec:estimation}

The mapping $\bm{\Theta} = \bm{F} (\bm{X})$ can be approximated by learning from historical pairs of weather report data and SDE parameters, denoted as $\{\bm{X}^{[k]},\bm{\Theta}^{[k]}\}_{k=1,\ldots,N}$. The weather report data are directly available from the local meteorological service; however, the historical SDE parameters are not directly available. Instead, they need to be estimated from historical PV data. 

Suppose historical PV generation data for an hour denoted by $\{\tilde{P}_{ih}\}_{i=1,\ldots,N_\mathrm{s}}$ with sampling period $h$ and size $N_\mathrm{s} = 3600/h$. Theoretically, the SDE parameters $\bm{\theta}$ in (\ref{eq:ito}) 
can be directly estimated by the maximal likelihood 
\cite{qiu2021nonintrusive}.
However, due to the relatively small sample size, e.g., 120 points per hour in this work, 
direct estimation may lead to numerical instability \cite{bibby1995martingale} and result in remarkable noise, which poses additional challenges in approximating $\bm{F}(\cdot)$. To alleviate this issue, a modified procedure is designed, where the drift and diffusion terms are estimated separately.

The first step is to identify the parameters of the diffusion term. The discrete version of the \emph{It\^{o}'s isometry} \cite{Pardoux2014Stochastic} yields
\begin{align}
   & \sum\nolimits_{i=1}^{N_\mathrm{s}} \Delta\tilde{P}_{ih}^2 \approx \sum\nolimits_{i=1}^{N_\mathrm{s}} h \sigma^2(\tilde{P}_{ih}; \bm{\theta}), \label{eq:itoisometry}
\end{align}
\noindent
where $\Delta\tilde{P}_{ih} \triangleq \tilde{P}_{(i+1)h} - \tilde{P}_{ih}$. Note that the parameters in the drift term, i.e., $a$ and $b$, do not appear in (\ref{eq:itoisometry}). Therefore, without knowing the drift term,
the parameters $c$, $d$, and $\beta$ of the diffusion term can be estimated first by
\begin{align}\hspace{-2pt}
   \max_{ c, d,\beta }\ L_{\sigma}(\bm{\theta}) = -\sum_{i=1}^{N_\mathrm{s}} \Big\Vert \Delta\tilde{P}_{ih}^2 - h \beta \big(\tilde{P}_{ih} - c\big) \big(d - \tilde{P}_{ih}\big) \Big\Vert^2.
\end{align}

The second step is to estimate $a$ and $b$ in the drift term. To alleviate the numerical stability issue, the \emph{martingale likelihood estimation} \cite{bibby1995martingale} is introduced, as follows:
\begin{align}  \hspace{-4pt}
\max_{a,b}\ L_{\mu}(\bm{\theta}) = \sum_{i=1}^{N_\mathrm{s}} \frac{\mu(\tilde{P}_{ih};\bm{\theta})}{\sigma^2(\tilde{P}_{ih})}  \left( \tilde{P}_{(i+1)h} - \mathbb{E} [ P_{(i+1)h} | \tilde{P}_{ih}] \right)
\end{align}
\noindent
where the conditional expectation on the right-hand side can be approximated based on \emph{It\^{o}'s lemma}, as follows:
\begin{align}
   \mathbb{E} & [ P_{(i+1)h} | \tilde{P}_{ih}] \approx \tilde{P}_{ih} + h \mu( \tilde{P}_{ih};\bm{\theta})  \\
   +  &  \frac{1}{2}h^2 \left[ \mu( \tilde{P}_{ih};\bm{\theta}) \frac{\partial\mu}{\partial P} (\tilde{P}_{ih};\bm{\theta})  +  \sigma^2( \tilde{P}_{ih}) \frac{\partial^2\mu}{\partial P^2} (\tilde{P}_{ih};\bm{\theta})\right]. \nonumber
\end{align}

Because different hours are mutually independent in parameter identification, the parameters for each hour can be estimated separately. Therefore, only a batch of low-dimensional programming problems needs to be solved. The overall procedure is computationally easy and efficient.

\subsection{Mapping Structure Design}
\label{sec:structure}

\begin{figure}[tb]
  \centering
  \includegraphics[width=3.45in]{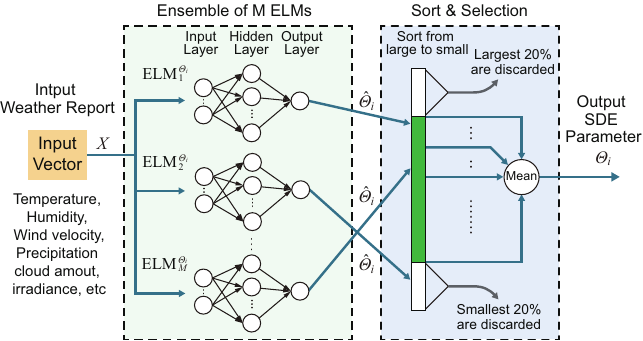}\vspace{-4pt}
\caption{Mapping structure 
 from the public weather data to the SDE parameter.}
  \label{fig:bagging}\vspace{-8pt}
\end{figure}


The structure for approximating the weather-to-parameter mapping $\bm{F}(\cdot)$ is designed based on the following factors.

\emph{1)} The size of training sample is limited. The public weather report data is typically resolutioned hourly. This means the valid size of training samples is relatively small. 
Therefore, although deep neural networks (DNNs) have excellent performance in scenario generation \cite{chen2018model} and forecast \cite{
khodayar2017rough},
considering its requirement for a large training set, a DNN-based structure seems unsuitable. 
See discussion in Section \ref{sec:dl}.

Because the single-layer extreme learning machine (ELM) \cite{huang2006extreme} has fewer parameters to tune, and it has reputed generalization ability and robust performance on small training sets, we choose to use the ELM to approximate $\bm{F}(\cdot)$.

\emph{2)} The training samples contain long-tail-distribution noise, which include the mismatch between the public weather report and the actual meteorological conditions at the PV station, and the numerical noise introduced in identifying the historical SDE parameters. Extreme weather such as typhoons in coastal areas also imposes difficulties in approximating $\bm{F}(\cdot)$.

To reduce the impact of noise and further enhance the robustness, the \emph{bootstrap aggregating} technique \cite{efron1994introduction}, also known as \emph{bagging}, is employed, where an ensemble of bootstrapped ELMs is trained instead of only one.

Overall, the structure for approximating the weather-to-parameter mapping is illustrated in Fig. \ref{fig:bagging}.

\subsection{Using Extreme Learning Machines to Map Public Weather Reports to SDE Parameters}
\label{sec:elm}

The ELM \cite{huang2006extreme} is a single-hidden-layer neural network with random input weights and biases. Denoting the number of hidden neurons as $K$, each ELM in this work is associated with one entry of the SDE parameter vector $\bm{\Theta}$, as 
\begin{align}
   \hat{\Theta}_i(\bm{X}) = \sum\nolimits_{j=1}^K v_j f (\bm{w}_j \bm{X} + b_j) \label{eq:elm}
\end{align}
where $\hat{\cdot}$ represents the approximation,
$\bm{w}_i$ is the $i$th row of the weight matrix $\bm{w}$; and $b_j$ is the $j$th entry of the bias vector $\bm{b}$. Each entry of $\bm{w}$ and $\bm{b}$ is drawn from an independent continuous random variable, specifically $\mathscr{N}(0,1)$ in this work. $v_j$ is the $j$th entry of the output weight vector $\bm{v}$; the sigmoid function selected for the activation $f(\cdot)$.

The ELM is trained by calculating the output weight $\bm{v}$, as
\begin{align}
    {\bm{v}} = \bm{H}^\dag  \left[ \Theta_i^{[1]}, \ldots, \Theta_i^{[N]} \right]^{\mathrm{T}} \label{eq:pseudoinverse}
\end{align}
\noindent
where $^\dag$ represents the pseudoinverse; $\bm{H}$ is the hidden output matrix with each entry defined by  $H_{jk} =  f (\bm{w}_k \bm{X}^{[j]} + b_k) $.


Because solving (\ref{eq:pseudoinverse}) only requires linear manipulations, the training process is extremely efficient \cite{huang2006extreme}. Furthermore, as an ELM only has $K$ parameters, it adapts well to the small training set available in this work.

Note that because the temporal patterns of PV power are already embedded in the SDE (\ref{eq:itohour}), here the ELM does not need to learn neither temporal nor probabilistic information. Hence, a standard ELM with RMSE loss function \cite{huang2006extreme} is sufficient.



\subsection{Bootstrapped ELM Ensembled to Address Numerical Noise and Extreme Weather Conditions}
\label{sec:bagging}

Because of extreme weather conditions 
and the numerical noise introduced in identifying the historical SDE parameters, the accuracy of a single ELM trained by the whole training set may not be satisfactory. To reduce these adverse impacts, the bootstrap aggregating technique \cite{efron1994introduction} is employed.

Specifically, from the original training set, $M$ bootstrapped training sets are resampled with replacement. Then, associated with each scaler parameter $\Theta_i$, an ensemble of $M$ ELMs is trained with the $M$ bootstrapped training sets.

Given input $\bm{X}$, the ELM ensemble provides $M$ scalar outputs. To remove extreme and abnormal values, the largest $20\%$ and the smallest $20\%$ of the outputs are discarded.
The mean value of the remaining $M\times60\%$ outputs is used as the parameter of the SDE model (\ref{eq:ito}). The selection process statistically alleviates the impact of numerical noise and extreme weather. The whole mapping structure is illustrated in Fig. \ref{fig:bagging}.

\section{Numerical Simulation Results}
\label{sec:case}

\subsection{Dataset Description }

The data\footnote{available at https://dx.doi.org/10.21227/1khg-8t55}
\label{sec:data} used in this study were collected in Macau covering 2018 and 2019. A $48$-hour weather report with a resolution of $1$ hour are publicly available at the Macao Meteorological and Geophysical Bureau website 
and include the temperature, humidity, atmospheric pressure, precipitation amount, wind direction and velocity, cloud amount, and irradiance.
The PV data with a $30$-second resolution were collected from a rooftop plant rated $P^{\text{rate}} = 2.9$ kW on the campus of the University of Macau.

After removing the days when the PV station was out of service for maintenance, we have $682$ days of valid data. Among them, $70\%$ ($N=479$) are randomly chosen as the training set, and the remaining $30\%$ are used as the testing set.

\begin{figure}[tb]
  \centering
    \includegraphics[scale=0.91]{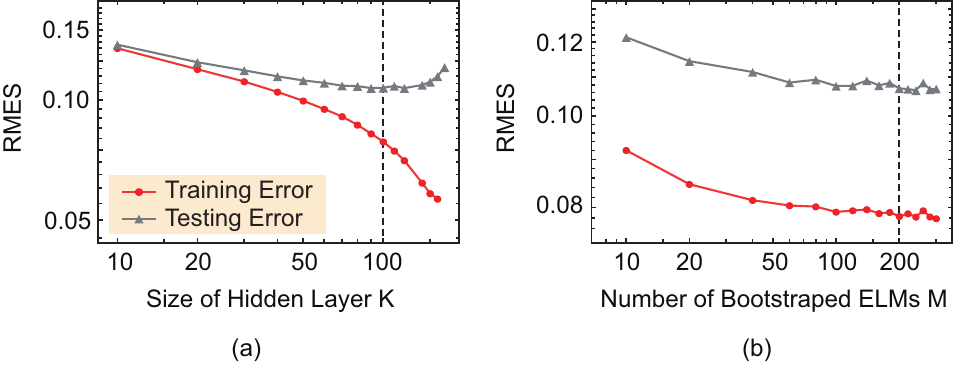}\vspace{-5pt}
    \caption{Training and testing RMSE of approximating the weather-to-parameter mapping (\ref{eq:map}) with different hyperparameters. (a) Size of the ELM hidden layer $K$. (b) Size of the bootstrapped ensemble $M$. }
  \label{fig:hyperpara}\vspace{-6pt}
\end{figure}

\subsection{Hyperparameter Selection}
\label{sec:hyperpara}

In the weather-to-parameter mapping designed in Section \ref{sec:map}, 
two hyperparameters are selected in advance, i.e., the size $K$ of the ELM hidden layer and the size $M$ of the ensemble.

We use the relative root-mean-square error (RMSE) to quantify the accuracy of the mapping.
The RMSEs on the training and testing sets with $M=200$ and different $K$ values are plotted in Fig. \ref{fig:hyperpara}(a). 
As a result, $K$ is selected as $100$ to minimize the testing error and to avoid either underfitting or overfitting.
Meanwhile, the influence of $M$ on the accuracy is plotted in Fig. \ref{fig:hyperpara}(b).
As a tradeoff between accuracy and computational cost, we choose $M=200$.

With $5$ parameters per hour, $12$ daytime hours per day and the ensemble size $200$, the total number of trained ELMs is $5\times 12\times 200 =12,000$. The overall training time is $33.8$ s on a desktop with an \emph{i7-8700} CPU. The RMSEs of the obtained weather-to-parameter mapping on the training and testing sets are $8.10\%$ and $11.79\%$, respectively.

Note that the RMSE of the weather-to-parameter mapping 
does not directly indicate the accuracy of the SDE model (\ref{eq:itohour}). To assess the accuracy of the proposed SDE model, we need to fill in the parameters and then compare simulated time series to the actual PV data, which is elaborated as follows.

\subsection{Performance Metrics for Time-Series Forecast}
\label{eq:metric}

Metrics for time-series forecast are employ to quantify the accuracy of the SDE model, as well as to facilitate comparison with state-of-the-art time-series forecast models, including:


\emph{a. Prediction Interval Coverage Probability (PICP)}: assessing the accuracy of the predictive interval, defined as the percentage of time the actual PV power stays within the predictive confidence interval; see \cite{mashlakov2021assessing}. The $90\%$-PICP is used.

\emph{b. Kullback-Leibler (K-L) Divergence}: measuring the difference between the predictive distribution and the empirical distribution of the actual data. This metric is widely used in distributionally robust optimization; see definition in \cite{chen2018distributionally}. Zero means two distributions are the same.

\emph{c. $\rho$-Risk}: quantifying the accuracy of a quantile $\rho$ of the predictive distribution; see definition in \cite{salinas2020deepar}. The $0.5$- and $0.9$-risk metrics are used, with respect to the median and tail values.

\emph{d. ND, and e. NRSME}: measuring the accuracy in terms of point forecast, respectively in the $1-$ and $2-$norms; see \cite{salinas2020deepar}. For probabilistic forecasts including the proposed SDE model, the median path of the Monte Carlo simulations is evaluated.

\emph{e. Autocorrelation Mismatch}, measuring the temporal correlation accuracy of the predictive time series. This is defined as the mean relative difference between the autocorrelation functions of the predictive time-series and the actual data. According to the decaying rate, a $3$-hour window is used.

Using these metrics, the performance of the models in terms of interval forecast, probabilistic forecast, and time-series forecast is comprehensively quantified.

\subsection{Accuracy of the Proposed Weather-Dependent SDE Model for Time-Series Forecast}
\label{sec:simu}

\begin{figure}[tb]
  \centering
    \includegraphics[scale=0.91]{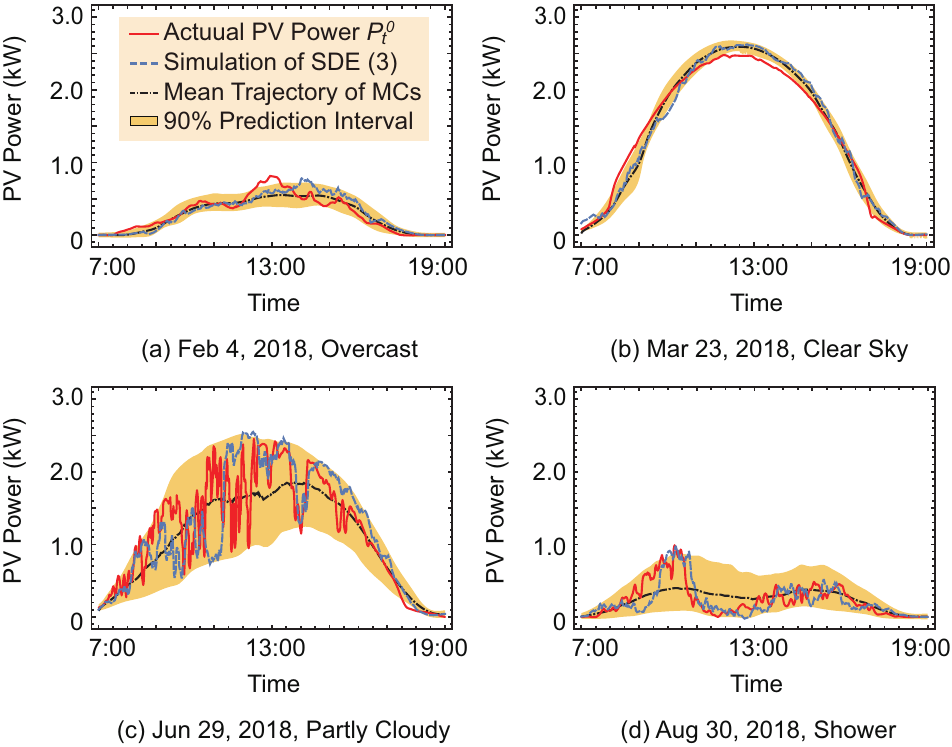}\vspace{-5pt}
\caption{Time-domain simulation of the proposed SDE model, the prediction interval, and the actual PV power under typical weather conditions.}
  \label{fig:sdeSim}\vspace{0pt}
\end{figure}

\begin{figure}[tb]
  \centering
  \includegraphics[scale=0.91]{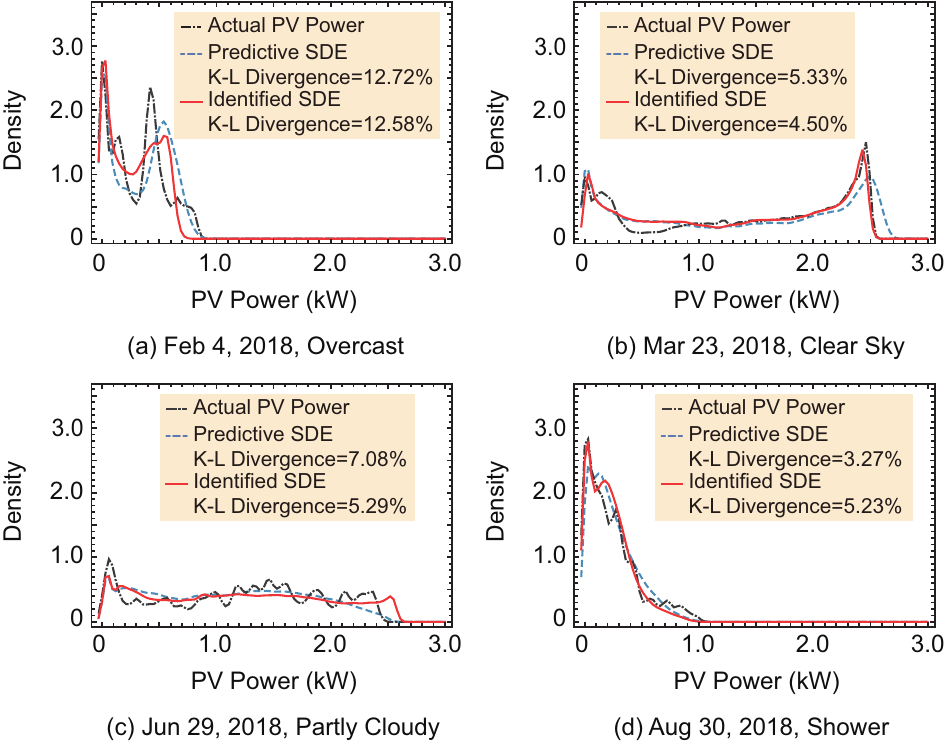} \vspace{-5pt}
  \caption{Densities of the proposed predictive SDE model, the identified SDE model, and the actual PV power under typical weather conditions.} 
  \label{fig:pdf}\vspace{-10pt}
\end{figure}

\begin{figure}[tb]
  \centering
    \includegraphics[scale=0.91]{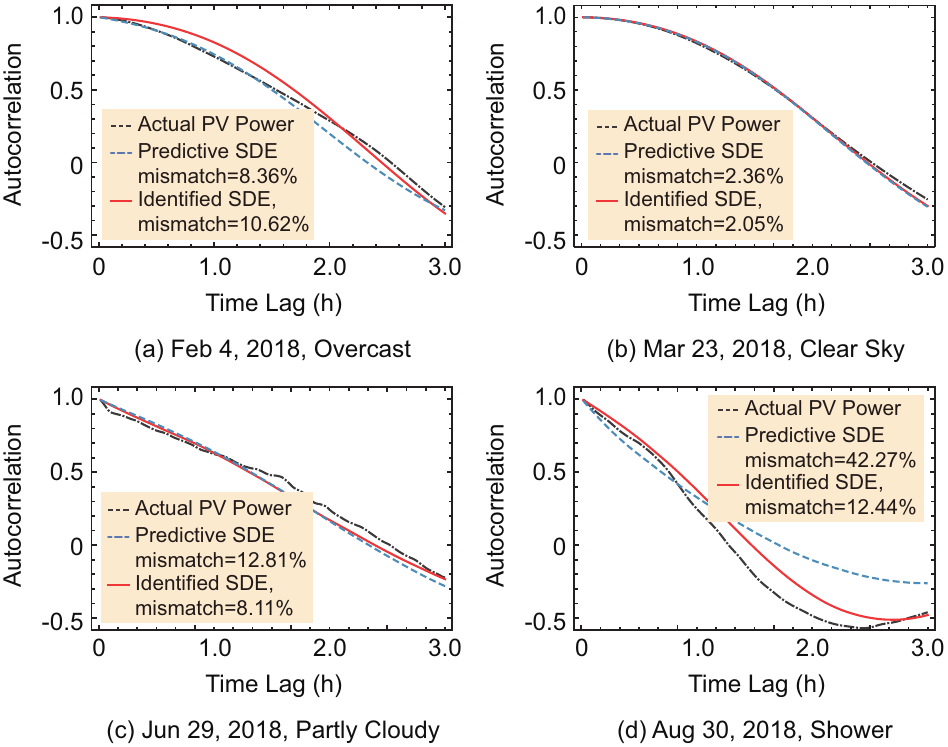} \vspace{-5pt}
    \caption{Time-series autocorrelation of the proposed predictive SDE model, the identified SDE model, and the actual data under typical weather conditions.}
  \label{fig:autoco}\vspace{-15pt}
\end{figure}

\subsubsection{Visual Validation}


Four days in the testing set with different weather are selected for visual validation.
Fig. \ref{fig:sdeSim} shows the simulation of the proposed predictive SDE model with parameters mapped from the weather report (simply referred to as the \emph{predictive SDE model} in the rest of the paper), compared with the real data and a non-predictive SDE model (parameters are directly identified from the PV data).
The mean trajectories and $90\%$-prediction intervals of the SDE models obtained by the Monte Carlo simulation (MCs) are shown in Fig. \ref{fig:sdeSim}.

Visually, the proposed predictive SDE model characterizes the weather-dependent uncertainty of the PV power very well. Different temporal patterns under different weather are clearly seen, and the prediction interval tightly fits the actual data. Additionally, thanks to the piecewise parameterization of the SDE as (\ref{eq:itohour}), the changes between different hours are clearly predicted.
Note that the SDE is a random process, hence, its simulation does not necessarily overlap with the actual data.

Another interesting phenomenon is that the mean simulation trajectory of the SDE model does not always lie in the exact middle of the prediction interval. This is because the probability distribution of PV
power is generally not centered, which is common in cloudy conditions.



\subsubsection{Statistical Validation}

In addition to the visual examination of the time-domain simulation, statistical similarity is also examined. 
For the four selected days, Fig. \ref{fig:pdf} shows the densities of the daily PV power provided by the predictive SDE model, the identified SDE model, and the actual PV data; and the time-series autocorrelation functions for each day are given in Fig. \ref{fig:autoco}. The K-L divergence and autocorrelation mismatch metrics are also labeled in the figures.


From the results, we know that the proposed predictive SDE model also provides a decent statistical forecast of the PV power. Both the predictive probability density and autocorrelation function pretty much resemble the actual PV generation data, just as the directly identified SDE model.

Note that in Figs. \ref{fig:pdf}(c) and \ref{fig:pdf}(d) the density function of the actual PV data fluctuates to some extent. This is attributed to sampling noise, as the actual PV data is only one sample of its underlying random process. Nevertheless, the proposed SDE model complies with the overall shape of the density function, avoiding overfitting these noisy fluctuations.





\subsection{Comparison with Deep Learning-Based Time Series Forecast Models}
\label{sec:dl}

Since the simulation result of the proposed SDE model provides time-series forecast of the PV power, it is natural to compare it with deep learning (DL)-based forecast models. For a comprehensive comparison, we select four latest DL-based models with different structures, covering recursive and non-recursive, and convolutional and non-convolutional, etc.

\begin{figure}[tb]
  \centering
  \includegraphics[width=3.48in]{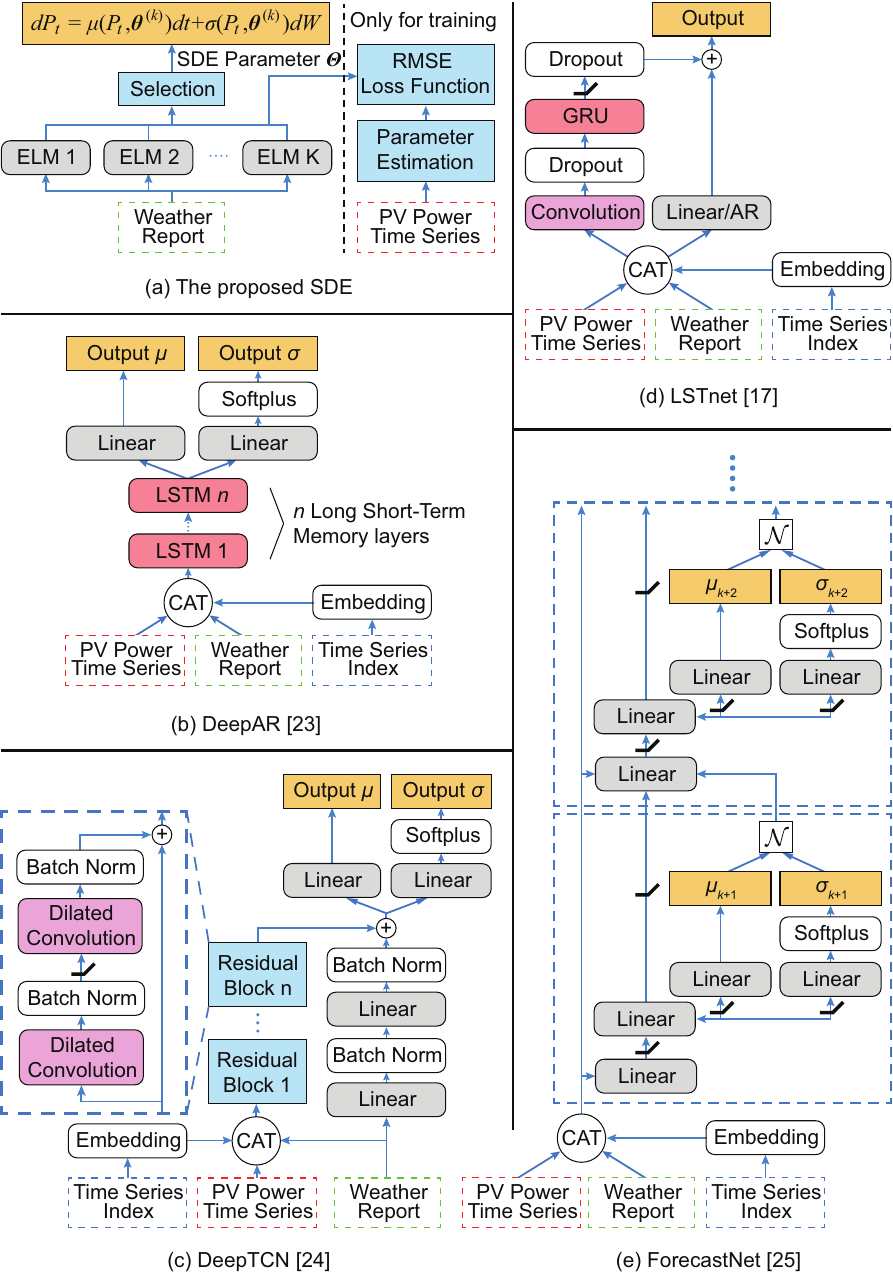} \vspace{-6pt}
  \caption{Diagrams of the proposed SDE model and latest deep learning-based models compared in Section \ref{sec:dl}. 
  }
  \label{fig:dl}\vspace{-12pt}
\end{figure}

The selection includes: 1) \emph{DeepAR} \cite{salinas2020deepar}, a long short-term memory-based deep recursive network. 
2) \emph{DeepTCN} \cite{chen2020probabilistic}, a deep residual convolutional network. 
3) \emph{LSTnet} \cite{lai2018modeling}, a hybrid of a linear autoregressive (AR) model and a convolutional recursive network 
that captures nonlinear patterns. 
4) \emph{ForecastNet} \cite{dabrowski2020forecastnet}, a purely feed-forward network without convolutional or recursive structure. According to the authors, 
its heterogeneity helps in capturing the non-homogeneous patterns of a non-stationary time series \cite{dabrowski2020forecastnet}. 




These DL-based models are illustrated in Fig. \ref{fig:dl} with the SDE-based model.
Because the resolution of the public weather report is $1$ hour, in training data the PV power is paired with the same weather report data for each hour. The hyperparameters of the DL-based models are tuned by a grid search.

Structurally, the proposed SDE-based model is the simplest.
As elaborated in Section \ref{sec:jacobi}, the temporal patterns of PV power are inherently embedded in the Jacobi diffusion process (\ref{eq:ito}).
Using only $5$ parameters, the SDE can recreate the temporal patterns of PV power for an hour under various weather conditions.
Hence, only a simple learning model is needed to establish the relation between the weather report data and these SDE parameters.
This remarkable reduces model complexity.

In contrast, the DL-based models use different neural network structures to recreate the temporal patterns. Unlike the proposed SDE-based model, no prior mathematical knowledge associated with the random process of PV power is exploited. Hence, their model structures are much more complex and there are more parameters to train.

The performance metrics of the proposed predictive SDE model and the DL-based forecast models on the testing set are compared in Table \ref{tab:perf}. For benchmarking, the SDE model directly identified from the PV data is also included. Except for PICP, other metrics are normalized by DeepAR (S=6) for clarity. Because LSTnet does not provide a probabilistic forecast, its PICP and $\rho$-risk metrics are not available.

Among the metrics, the proposed SDE-based model performs among the best in $90\%$-PICP, K-L Divergence, $0.9$-risk, ND, and Autocorrelation Mismatch. Although LSTnet also performs among the best in ND, and ForecastNet in $0.5$- and $0.9$-risks and NRSME, the proposed model has the largest number of the best metrics.
Especially, the Autocorrelation Mismatch of the SDE-based model is much better than all of the DL-based models. This is simply because the Jacobi diffusion process (\ref{eq:ito}) relies on less information  than the neural network-based models to recreates the temporal patterns of PV power, as explained in this section before.
As a result, 
when only the public weather report is available,
the SDE-based model outperforms the selected DL-based models in PV power forecast.

\begin{table*}[tb]\scriptsize
  \renewcommand{\arraystretch}{1.42}
  \centering
  \caption{Time-Series Forecast Performance Metrics of the Proposed SDE Model and a Selection of Deep Learning-Based Models}\vspace{-4.5pt}
  \label{tab:perf}
  \centering
  \begin{tabular}{cccccccc}
  \hline \hline
  \multirow{3}{*}{Model}        & \multicolumn{7}{c}{Performance Metrics} \\
  \cline{2-8}                   & \multirow{2}{*}{\tabincell{c}{ $90\%$-Prediction Interval \\  Coverage Probability (PICP)}} & \multicolumn{3}{c}{Probability Distribution} & \multicolumn{2}{c}{Point Forecast} & \multirow{2}{*}{\tabincell{c}{Autocorrelation \\Mismatch}} \\
  \cline{3-7}                          &                     &  \tabincell{c}{K-L Divergence} & 0.5-Risk  & 0.9-Risk  & ND  & NRMSE  \\\hline
  DeepAR \cite{salinas2020deepar} (S=24)  & $82.9\%$         & $1.593$           & $0.915$         & $0.449$         & $0.940$          & $0.944$          & $0.995$          \\
  DeepAR \cite{salinas2020deepar} (S=12)  & $84.5\%$         & $1.345$           & $1.044$         & $0.902$         & $1.049$          & $1.032$          & $1.045$         \\
  DeepAR \cite{salinas2020deepar} (S=6)   & $88.1\%$         & $1$               & $1$             & $1$             & $1$              & $1$              & $1$   \\
  DeepTCN \cite{chen2020probabilistic}    & $78.1\%$         & $2.660$           & $2.199$         & $0.572$         & $1.528$          & $1.658$          & $1.162$         \\
  LSTnet \cite{lai2018modeling}           & N.A.             & $1.415$           & N.A.            & N.A.            & $\bm{0.915}$     & $0.893$          & $1.354$         \\
  ForecastNet \cite{dabrowski2020forecastnet}    & $92.8\%$         & $1.289$           & $\bm{0.459}$    & $\bm{0.285}$    & $0.942$          & $\bm{0.876}$     & $0.908$         \\
  Proposed Predictive SDE                 & $\bm{89.6\%}$    & $\bm{0.988}$      & $0.702$         & $\bm{0.286}$    & $\bm{0.918}$     & $0.927$          & $\bm{0.753}$         \\ \hline
  SDE Directly Identified from Data       & $95.3\%$         & $0.590$           & $0.445$         & $0.197$         & $0.721$          & $0.806$          & $0.695$   \\
  \hline \hline
  \end{tabular} \\\vspace{-10pt}
\end{table*}

\subsection{Discussion on the Application in Uncertainty Quantification and Stochastic Optimization }
\label{sec:app}

DL-based PV forecast models cannot be analytically embedded into the uncertainty quantification (UQ) or stochastic optimization (SO) model of the electrical energy system. They only provide samples (called scenarios) of PV generation. Hence, the UQ requires the MCs, and the SO relies on scenario-based optimization, both very time-consuming.

In contrast, the SDE model of PV power provided by this work can be analytically incorporated into the differential equation model of the power system in stochastic calculus \cite{wang2015long,Chen2018Stochastic,qiu2021nonintrusive,qin2019stochastic}. Then, associated partial differential equations (including the Fokker-Planck equation and Feynman-Kac formula) and decomposition techniques (e.g., Karhunen-Lo\`{e}ve expansion and polynomial chaos) can be utilized to deal with it.
This enables much more efficient UQ and SO approaches.
According to 
\cite{Chen2018Stochastic,qiu2021nonintrusive,qiu2020fast}, SDE-based UQ is two orders of magnitude faster than the MCs; and according to 
\cite{chen2019optimal} and \cite{Chen2020OptimalC}, SDE-based SO can be $50$ times faster than the scenario-based method.
Due to space limit, here we will not go further. Interested readers are referred to these articles.

\section{Conclusions}
\label{sec:mcs}

An SDE-based random process model for PV power is proposed in this article. The model only relies on public weather reports instead of the high-resolution NWPs. A parameterized Jacobi diffusion process is constructed to recreate the temporal patterns of PV power. An ELM ensemble is established to map the weather report data to the parameters of the Jacobi diffusion process, which accurately captures the time-varying and weather-dependent uncertainty of PV power.

Compared to a selection of state-of-the-art deep learning-based forecast models, the proposed SDE-based model has a much simpler structure but is more accurate in forecasting the PV power when only public weather reports are available. The proposed model also exclusively enables the latest SDE-based uncertainty quantification and stochastic optimization techniques, which are much more efficient than the traditional scenario-based approaches.


The current SDE model can be extended to multiple dimensions. Hence, generalizing the proposed model to multiple correlated PV plants is one of the future studies.

\bibliographystyle{IEEEtran}
\bibliography{IEEEabrv,SDEPV}

\end{document}